\documentclass[pdflatex,sn-mathphys-num]{sn-jnl}

\usepackage{graphicx}%
\usepackage{multirow}%
\usepackage{amsmath,amssymb,amsfonts}%
\usepackage{amsthm}%
\usepackage{mathrsfs}%
\usepackage[title]{appendix}%
\usepackage{xcolor}%
\usepackage{textcomp}%
\usepackage{manyfoot}%
\usepackage{booktabs}%
\usepackage{algorithm}%
\usepackage{algorithmicx}%
\usepackage{algpseudocode}%
\usepackage{listings}%
\usepackage{makecell}%
\usepackage{graphicx}%
\usepackage{adjustbox}%

\theoremstyle{thmstyleone}%
\theoremstyle{thmstyletwo}%

\theoremstyle{thmstylethree}%

\begin{document}

\title[Article Title]{Cesarean Scar Defect Segmentation in Transvaginal Ultrasound Images: a Dataset and Benchmark}


\author[1]{\fnm{Yuan} \sur{Tian}}\email{tianyuan\_ipmch@sjtu.edu.cn}
\equalcont{These authors contributed equally to this work.}

\author[2,3]{\fnm{Yue} \sur{Li}}\email{yue.li3@nottingham.edu.cn, scxyl3@nottingham.ac.uk}
\equalcont{These authors contributed equally to this work.}

\author[1]{\fnm{Wei} \sur{Xia}}\email{xiaweiyz@163.com}
\equalcont{These authors contributed equally to this work.}

\author[4]{\fnm{Tianyu} \sur{Xu}}\email{tix010@ucsd.edu}

\author*[1]{\fnm{Jian} \sur{Zhang}}\email{zhangjian\_ipmch@sjtu.edu.cn}

\author*[5]{\fnm{Liye} \sur{Shi}}\email{timmy\_ye@163.com}

\author[2,3]{\fnm{Jing} \sur{Liu}}\email{jing.liu@nottingham.edu.cn, scxjl7@nottingham.ac.uk}

\author[1]{\fnm{Yang} \sur{Wang}}\email{wangyang0922@foxmail.com}

\author[6]{\fnm{Ming} \sur{Liu}}\email{liuming198904@sjtu.edu.cn}

\author[2,3]{\fnm{Qing} \sur{Xu}}\email{qing.xu@nottingham.edu.cn, scxqx1@nottingham.ac.uk}

\author[2]{\fnm{Yixuan} \sur{Zhang}}\email{scxyz8@nottingham.edu.cn}

\author[7]{\fnm{Maggie M.} \sur{He}}\email{Maggie.he@health.qld.gov.au}

\author*[2,3]{\fnm{Xiangjian} \sur{He}}\email{sean.he@nottingham.edu.cn, sean.he@nottingham.ac.uk}

\affil*[1]{\orgdiv{Department of Obstetrics and Gynecology}, \orgname{International Peace Maternity and Child Health Hospital affiliated to Shanghai Jiao Tong University School of Medicine}, \orgaddress{\street{910 Hengshan Road}, \city{Xuhui District}, \postcode{200030}, \state{Shanghai}, \country{China}}}

\affil*[2]{\orgdiv{School of Computer Science}, \orgname{University of Nottingham Ningbo China}, \orgaddress{\street{199 Taikang East Road}, \city{Ningbo}, \postcode{315100}, \state{Zhejiang}, \country{China}}}

\affil*[3]{\orgdiv{School of Computer Science}, \orgname{University of Nottingham}, \orgaddress{\street{University Park}, \city{Nottingham}, \postcode{NG7 2RD}, \country{UK}}}

\affil[4]{\orgdiv{Department of Computer Science and Engineering}, \orgname{University of California, San Diego}, \orgaddress{\street{9500 Gilman Drive}, \city{La Jolla}, \postcode{92093}, \state{CA}, \country{USA}}}

\affil*[5]{\orgdiv{Department of Ultrasound}, \orgname{International Peace Maternity and Child Health Hospital affiliated to Shanghai Jiao Tong University School of Medicine}, \orgaddress{\street{910 Hengshan Road}, \city{Xuhui District}, \postcode{200030}, \state{Shanghai}, \country{China}}}

\affil[6]{\orgdiv{School of Electronic Information and Electrical Engineering}, \orgname{Shanghai Jiao Tong University}, \orgaddress{\street{800 Dongchuan Road}, \city{Minhang District}, \postcode{200240}, \state{Shanghai}, \country{China}}}

\affil[7]{\orgdiv{Department of Cardiology}, \orgname{Gold Coast University Hospital}, \orgaddress{\street{1 Hospital Boulevard Southport}, \city{Gold Coast}, \postcode{4215}, \state{QLD}, \country{Australia}}}


\abstract{Cesarean Scar Defect (CSD) is one of the most prevalent complications following cesarean delivery. Transvaginal ultrasonography is widely used for primary CSD screening. Accurate determination of CSD outline and dimensions is crucial for treatment. However, CSDs are frequently overlooked by sonographers due to small size and irregular morphology, suboptimal image quality, and limited clinical awareness in resource-constrained settings. Despite artificial intelligence advances in medical imaging, no public dataset exists for transvaginal ultrasound CSD segmentation. To address this gap, we present a comprehensive CSD dataset comprising 1,111 images and 16 videos, yielding 501 positive samples with confirmed CSD and precise pixel-level manual annotations. Annotations are performed following standardized clinical guidelines through collaboration between experienced sonographers and trained PhD students. This work provides high-quality benchmark resources for advancing medical image segmentation algorithms and promoting clinical innovation. Ultimately, improved CSD diagnosis and subsequent treatment strategies can enhance the quality of life in women of reproductive age, representing significant value for both medical research and clinical practice.}




\maketitle

\section*{Background \& Summary}\label{sec1}

\begin{figure}[!t]
    \centering
    \includegraphics[width=1\linewidth]{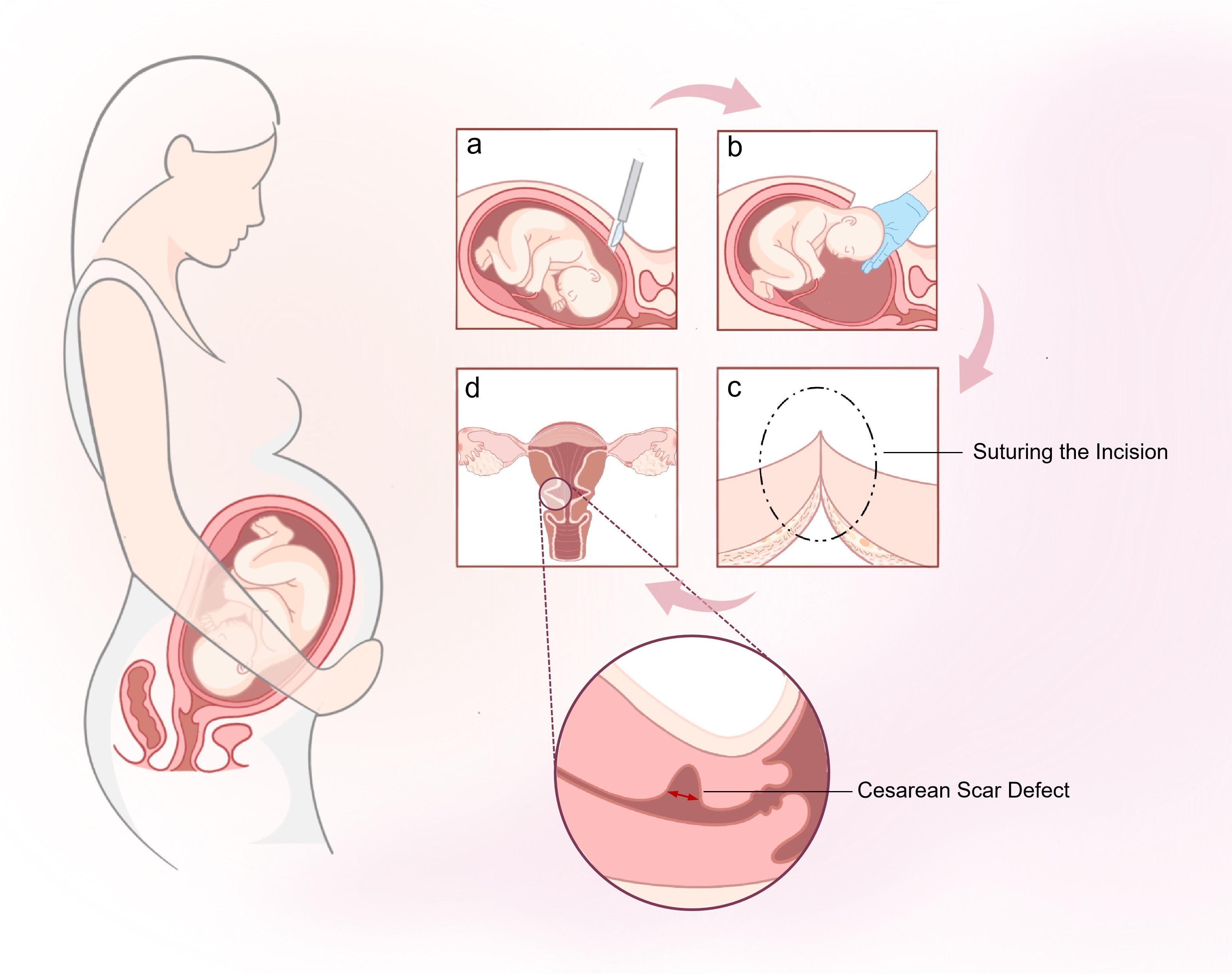}
    \caption{Schematic illustration of cesarean delivery and Cesarean Scar Defect (CSD) formation. This figure illustrates the main steps of cesarean delivery and the potential formation of uterine scar defects postoperatively. (\textbf{a}) Hysterotomy of the lower uterine segment. (\textbf{b}) Delivery of the fetus through the uterine incision. (\textbf{c}) Suturing the hysterotomy incision. (\textbf{d}) During the postoperative healing process, incomplete wound healing at the incision site may result in CSD.}
    \label{fig:1}
\end{figure}

Cesarean Scar Defect (CSD) is a common long-term complication of cesarean delivery caused by incomplete healing of the uterine myometrium. As shown in Figure \ref{fig:1}, CSD develops at the site of the uterine incision after cesarean delivery, where inadequate wound healing results in a persistent niche or indentation in the anterior uterine wall. According to the official definition, CSD is defined as \lq\lq a myometrial defect of at least 2mm depth at the cesarean scar site assessed by transvaginal ultrasound" \cite{pan2019prevalence}. Defects are classified as significant when the residual myometrial thickness (RMT) is less than 3mm \cite{meuleman2023definition}. These anatomical defects are present in approximately 60\% of women with a history of cesarean delivery, of whom 30-40\% develop symptomatic disease affecting daily functioning, including postmenstrual spotting, dysmenorrhea, chronic pelvic pain, or secondary infertility \cite{zhang2023comparing}. In addition, women of childbearing age with fertility desires face elevated risks during pregnancy and delivery when there is CSD. Serious obstetric complications, including Cesarean Scar Pregnancy (CSP), placenta accreta, placenta previa overlying the scar, scar dehiscence, uterine rupture, and postpartum hemorrhage, pose significant threats to both maternal and fetal well-being \cite{cali2018outcome, he2023fertility}. Therefore, early and accurate detection of CSD provides a critical window for timely risk stratification, clinical surveillance, and preventive intervention.

Current diagnostic methods for CSD primarily include MRI, transvaginal ultrasound (TVUS), saline infusion sonohysterography (SIS), hysterosalpingography and hysteroscopy \cite{budny2021uterine}. However, each of these approaches has inherent limitations that restrict their widespread or routine clinical use. MRI, while capable of providing high-resolution anatomical detail of the cesarean scar niche, is limited by high cost, restricted availability, prolonged examination time, and its inherently slice-based, non-continuous imaging nature, which constrains real-time and dynamic assessment \cite{bij2014prevalence, marotta2013laparoscopic}. In contrast, SIS, hysterosalpingography and hysteroscopy require intrauterine instrumentation or cavity manipulation, 
which may cause discomfort, increase the risk of infection or bleeding, and raise patient concerns regarding invasiveness and tolerability.

TVUS, owing to its low cost, wide availability, noninvasiveness and convenience, is the most commonly used first-line imaging modality for screening  CSD. However, reliable identification of CSD using conventional ultrasound remains challenging. CSD lesions are typically small, anatomically confined to the cervico-isthmic region posterior to the bladder, and morphologically irregular, and may be obscured by coexisting structures such as scar tissue, endometrial polyps or Nabothian cysts.   Moreover, ultrasound diagnosis is highly operator-dependent, resulting in substantial interobserver variability, particularly in primary care and general hospital settings where clinical experience with CSD is limited.  As CSD has only recently gained widespread recognition and standardized diagnostic criteria, insufficient awareness further contributes to missed and incorrect diagnoses.  Previous literature has shown that conventional diagnostic accuracy for CSD is only 24.0-69.1\% \cite{bij2011ultrasound, baranov2016assessment, pan2019prevalence}. These limitations highlight the need for more accurate, standardized, and reproducible ultrasound-based diagnostic approaches for CSD.

Artificial Intelligence, especially Deep Learning, has achieved multiple essential breakthroughs in medical image recognition tasks in recent years. Compared to imaging experts, deep learning algorithms perform better in the detection and evaluation of various diseases \cite{li2025uniultra}, such as breast diseases \cite{shen2021artificial}, thyroid diseases \cite{wang2020automatic} and ovarian tumors \cite{gao2022deep}. Jin et al. \cite{jin2022accuracy} studied multiple U-Net-based automatic segmentation models for cervical cancer transvaginal ultrasound images and their impact on radiomic features. It proves that it is feasible and reliable to use the regional features extracted by automatic segmentation to conduct the next step of the research.

Currently, no publicly available dataset exists for transvaginal ultrasound segmentation in Cesarean Scar Defect. To address this gap, we introduce the first comprehensive CSD dataset comprising 501 high-quality positive samples with fine pixel-level manual annotations, derived from 1,111 images and 16 videos collected during clinical examinations, and provide a benchmark for CSD segmentation tasks. The public release of this dataset offers multiple benefits: enhancing the medical community's understanding of CSD characteristics, supporting clinicians in developing more precise treatment strategies, and improving reproductive health outcomes for post-cesarean women. Additionally, this work establishes CSD segmentation as a novel task within medical image segmentation, opening new research avenues for gynecological ultrasound analysis. The dataset can reveal quantitative features and subtle patterns that may escape human visual detection, thereby enhancing diagnostic efficiency through automated analysis. Therefore, the publicly available CSD dataset provides essential resources for early CSD detection and treatment plan development, advances the broader field of medical image analysis and holds significant clinical application value for safeguarding women's reproductive health after cesarean delivery. 

\section*{Methods}\label{sec2}

\begin{figure}[!t]
    \centering
    \includegraphics[width=1\linewidth]{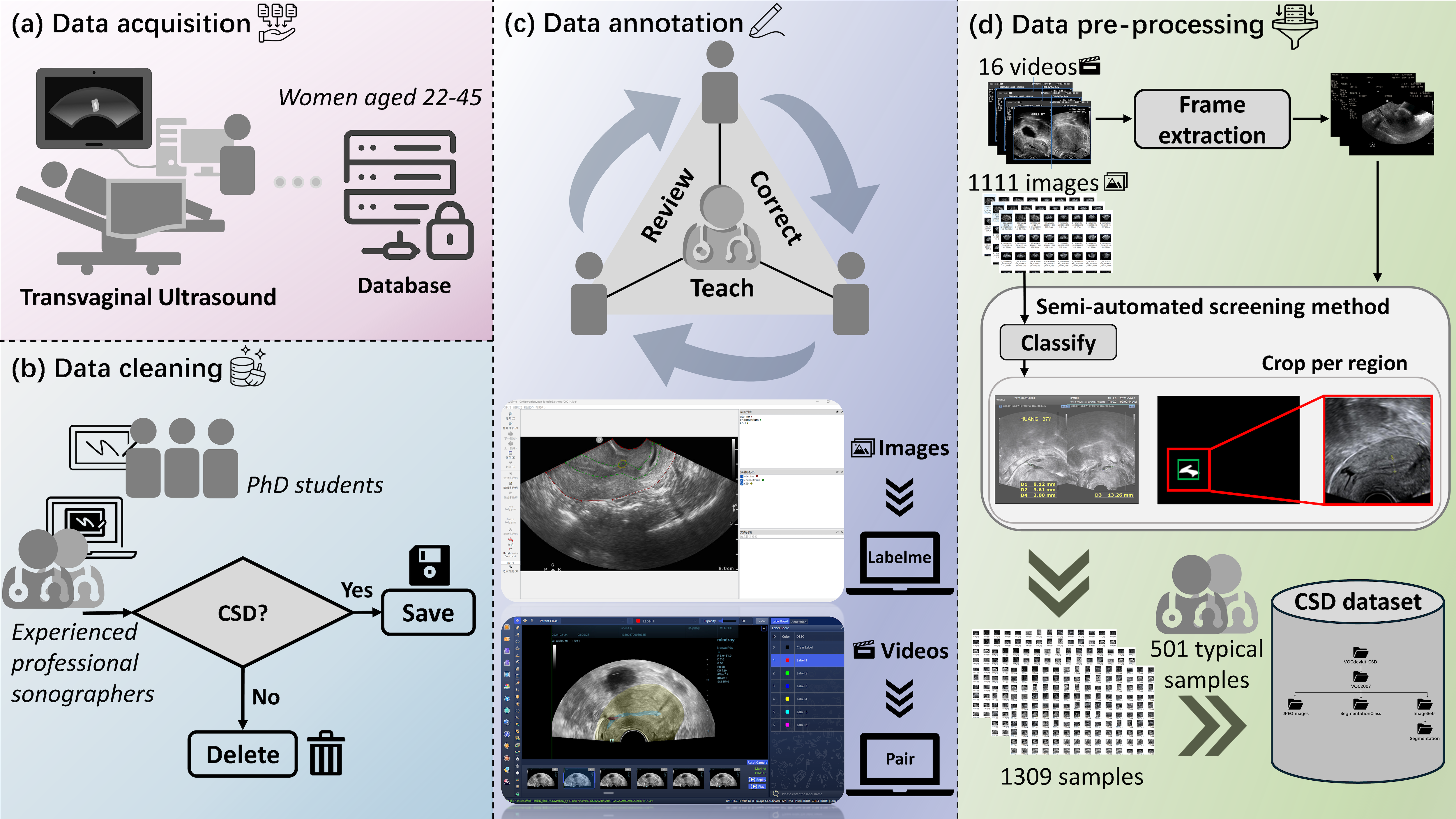}
    \caption{Overall workflow for collecting the Cesarean Scar Defect dataset. (\textbf{a}) Women aged 22-45 years underwent transvaginal ultrasound examinations, and the data was stored in the hospital database. (\textbf{b}) Experienced professional sonographers guided PhD students in determining whether the patient data exhibits CSD. (\textbf{c}) Sonographers taught PhD students how to annotate CSD regions. Each sample was annotated after discussion between two PhD students, and then the sonographer reviewed and modified the results. Image data was annotated using Labelme, while video data was annotated using Pair. (\textbf{d}) 16 videos and 1,111 images were cropped and anonymized using a semi-automatic screening method, and 501 positive samples were finally screened out to establish the CSD dataset.}
    \label{fig:2}
\end{figure}

\subsection*{Data Acquisition and Cleaning}
This single-center retrospective cohort study was approved by the Ethics Committee of the International Peace Maternity and Child Health Hospital (IPMCH), affiliated with Shanghai Jiao Tong University School of Medicine (Approval No. GKLW-A-2024-101-01) and University of Nottingham Ningbo China (Approval No. FOSE-202425-015). Given the use of fully anonymized medical imaging data, informed consent was waived. Patient inclusion criteria required all subjects to meet the following conditions: ultrasound diagnosis of significant uterine diverticulum (diverticulum depth $\geq$50\% of myometrial thickness, residual myometrial thickness $\leq$3mm), secondary infertility (failure to conceive for more than one year without contraception or recurrent in vitro fertilization implantation failure), and active fertility desire.

Figure \ref{fig:2} shows the overall workflow for collecting the CSD dataset. The CSD dataset was derived from patients treated at IPMCH, encompassing women aged 22-45 years who underwent transvaginal ultrasound examination. Raw data were acquired using standardized ultrasound equipment including Philips iU-22, Samsung W10, and Mindray Resona R9, yielding 1,111 static images and 16 dynamic video sequences without inherent spacing information. All clinical data from transvaginal ultrasound examinations were systematically archived in a secure, HIPAA-compliant database following institutional data management protocols.

A rigorous quality control process was implemented involving a multidisciplinary team of experienced professional sonographers and trained PhD students who independently reviewed all collected data. Images and videos demonstrating typical CSD morphological characteristics were retained, while those with insufficient image quality, suboptimal visualization, or absence of clear CSD features were systematically excluded from the dataset to ensure data integrity and clinical relevance.

\subsection*{Data Annotation}
A collaborative annotation framework was established incorporating multiple expert sonographers and trained PhD students through an iterative process of standardized teaching, systematic review, and quality correction to ensure inter-observer consistency and annotation accuracy. Both static ultrasound images and dynamic video sequences were processed using specialized medical annotation software, specifically Labelme for static image segmentation and Pair for video sequence annotation, ensuring precise delineation and pixel-level masking of CSD regions of interest. All annotations were performed according to established clinical guidelines and standardized protocols, with each annotation undergoing multi-level validation to maintain dataset quality and clinical applicability.

\begin{figure}[!t]
    \centering
    \includegraphics[width=1\linewidth]{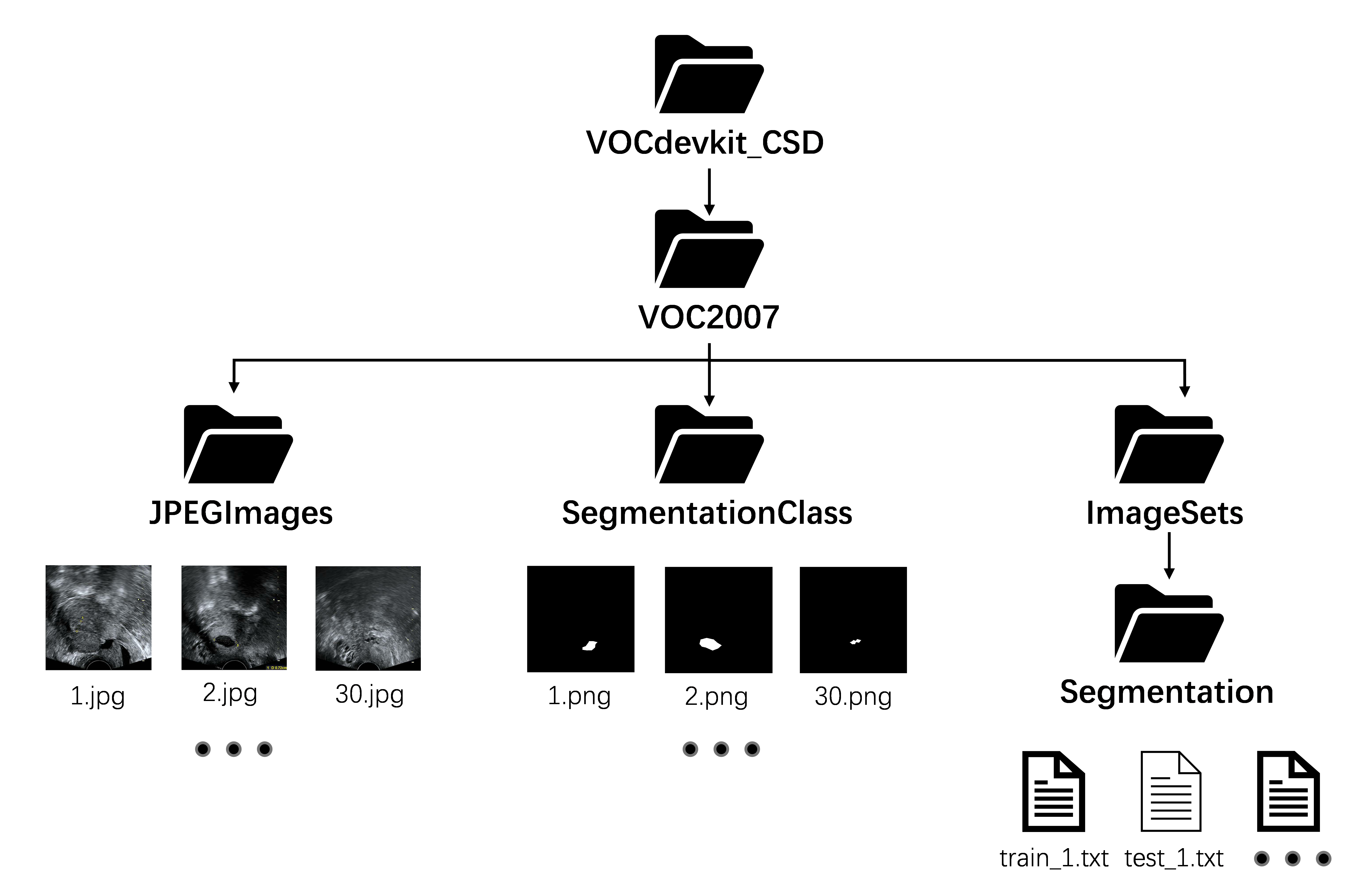}
    \caption{File structure.}
    \label{fig:structure}
\end{figure}

\subsection*{Data Pre-processing}

Firstly, we extracted every 10 frames from 16 labeled ultrasound videos. The overall redundancy of the data was overcome by eliminating the similarity between adjacent frames. Since there was a substantial amount of residual digital information from the ultrasound equipment around all images and video frames, a semi-automatic cropping algorithm was used to process all the labeled images into cropped shapes that focus on the region of interest (ROI). Then, the sonographers selected 501 typical samples from 1309 samples. Finally, those samples constituted the CSD dataset.

\section*{Data Records}\label{sec3}

All data are available on \href{https://kaggle.com/datasets/d751e364b67a71784d1eb6d0b543e6b7b46dcde03f5d227359110a204b37085b}{CSD Dataset}.

As shown in Figure \ref{fig:structure}, the dataset is organized by following the Pascal VOC format with a hierarchical directory structure. The root directory VOCdevkit\_CSD contains a subdirectory named VOC2007, which comprises three main folders. The JPEGImages folder stores all original images in JPEG format, with each file named using a numerical identifier (e.g., 1.jpg, 2.jpg). The corresponding pixel-wise segmentation masks are stored in the SegmentationClass folder in PNG format, where each mask shares the same filename as its corresponding original image to ensure a one-to-one correspondence. The ImageSets folder contains a subfolder named Segmentation, which includes text files (e.g., train\_1.txt, test\_1.txt) that specify the data splits by listing the numerical identifiers of images assigned to training and testing subsets, respectively. This standardized structure ensures compatibility with widely used deep learning frameworks and facilitates seamless integration into existing semantic segmentation pipelines.

\begin{table}[!t]
    \centering
    \caption{Five-fold cross-validation results (mean ± std).}
    
    \small  
    \begin{tabular}{l|cccccc}
    \hline
    \textbf{Method} &  \textbf{Dice} & \textbf{Precision} &
    \textbf{Recall}  & \textbf{IoU} & \textbf{HD95}\\
    \hline
    UNet \cite{ronneberger2015u}  & 74.79 ± 2.31  & 84.77 ± 2.25 & 83.68 ± 2.30 & 71.17 ± 2.14 & 8.78 ± 2.55 \\
    DeeplabV3+ \cite{chen2018encoder} & 75.92 ± 2.99  & 85.87 ± 4.38 & 83.93 ± 1.81 & 72.12 ± 3.06 & 7.78 ± 1.37 \\
    GCNet \cite{cao2019gcnet} & 74.98 ± 4.71  & 87.77 ± 7.67 & 81.72 ± 4.30  & 71.36 ± 4.86 & 6.60 ± 1.90 \\
    Swin-UNet \cite{cao2022swin} & 74.03 ± 2.34  & 87.57 ± 2.09 & 78.77 ± 2.23  & 70.44 ± 2.17 & 7.78 ± 2.07 \\
    
    \hline
    \end{tabular}
    
    \label{tab:1}
\end{table}

\begin{figure}[!t]
    \centering
    \includegraphics[width=1\linewidth]{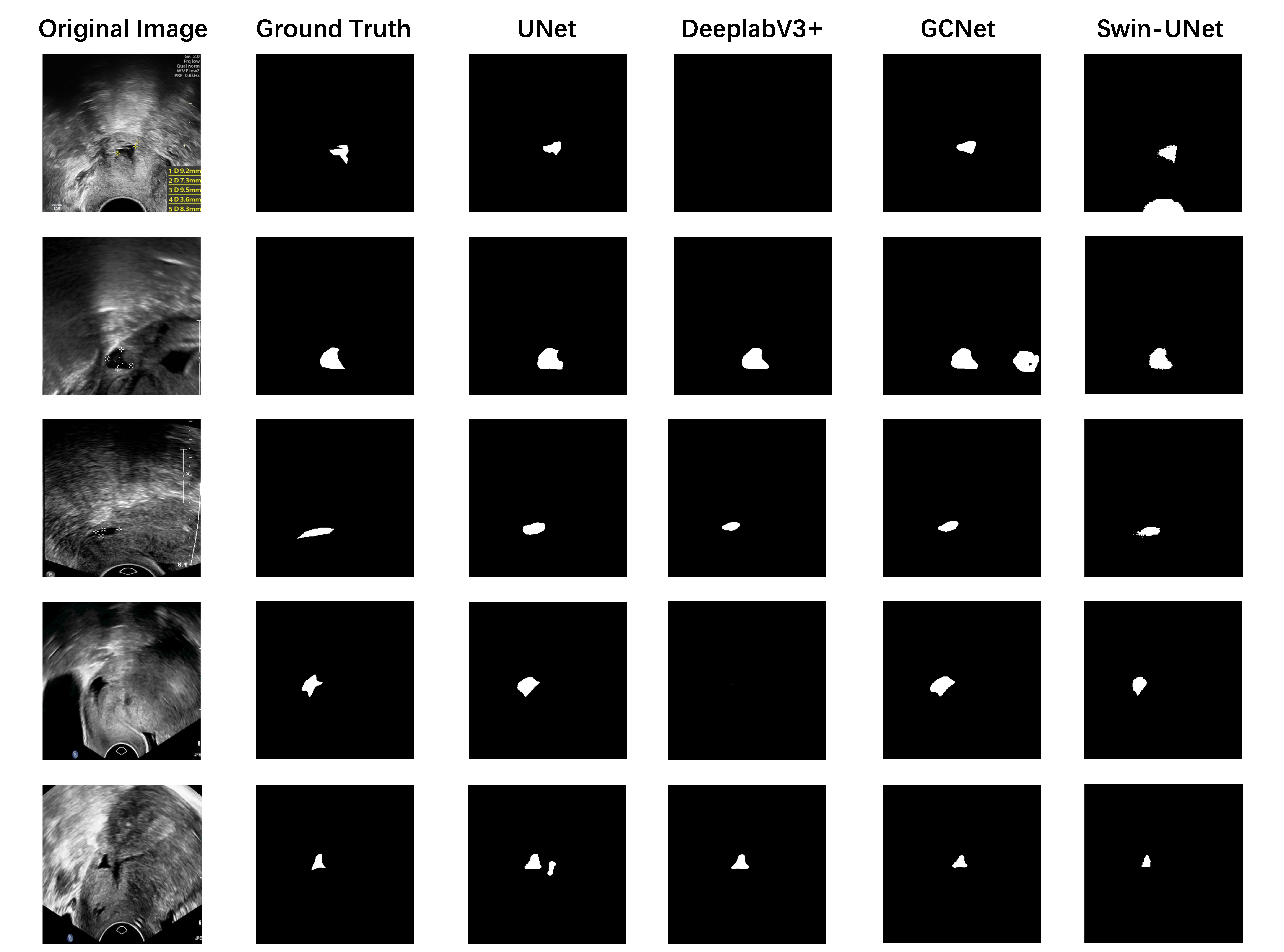}
    \caption{Visualization.}
    \label{fig:vis}
\end{figure}

\section*{Technical Validation}\label{sec4}

To verify the clinical relevance and segmentation quality of the CSD dataset, we benchmarked the annotations against four state-of-the-art medical image segmentation architectures under a rigorous five-fold cross-validation protocol. All 501 positively labeled images were used. Dice, Precision, Recall, Intersection-over-Union (IoU) and 95-percentile Hausdorff Distance (HD95) were computed after the predicted mask was aligned with the expert contour that served as ground truth. Mean ± standard deviation across the five folds is reported in Table \ref{tab:1}.

Across the four models, the mean Dice coefficient ranges from 74.03\% to 75.92\%, while IoU spans 70.44\%–72.12\%. The highest Dice (75.92 ± 2.99\%) and IoU (72.12 ± 3.06\%) are achieved by DeepLabV3+, and followed closely by GCNet (74.98 ± 4.71\% Dice). At the dataset level, no model exhibits systematically low Recall (minimum fold-mean 78\%), indicating that overall lesion omission is not a dominant failure mode. Nevertheless, isolated cases of incomplete coverage still occur. Precision is more variable (78\%–88\%), reflecting the challenge of the thin, irregular CSD morphology that occasionally confuses edge pixels with background. HD95 distances are below 9 voxels for every model (best 6.60 ± 1.90 mm for GCNet), confirming that the predicted boundaries remain within a clinically acceptable margin from the expert trace. The low standard deviations demonstrate that the performance is stable across folds. Therefore, the dataset size is sufficient to support robust training.

Taken together, these results validate two key points that: (i) the manual delineations are consistent enough to allow top-tier segmentation models to converge to clinically meaningful solutions, and (ii) the dataset poses a non-trivial but solvable challenge, leaving ample room for future algorithmic innovation.

\section*{Data Availability}\label{sec5}

The complete dataset will be publicly available for download in the Kaggle repository once the paper is accepted. Users can conduct further meaningful research based on this dataset to advance the development of medical and AI research.


\section*{Code availability}\label{sec6}

No novel code was used in the construction of the CSD dataset.

\section*{Acknowledgements}\label{sec7}

This work is supported by the Yongjiang Technology Innovation Project (2022A-097-G), Shanghai Jiaotong University Medical–Engineering Interdisciplinary Project (YG2026ZD33, YG2025QNA12, YG2024QNA58) and Shanghai Municipal Science and Technology Commission Medical Innovation Research Special Project (23Y11909100).

\section*{Author contributions}\label{sec8}

All the authors have reviewed the manuscript. Yuan Tian contributed through data collection and data annotation. Yue Li conceptualized the study, designed the experiments, performed data preprocessing, and was responsible for dataset construction, management, visualization, and drafting the manuscript. Wei Xia assisted with data acquisition and reviewed the manuscript. Tianyu Xu implemented the experiments, organized the data, and contributed to visualization. Liye Shi supervised the data collection and annotation processes. Jing Liu contributed to the generation of figures and visualization. Yang Wang participated in data acquisition. Ming Liu contributed to the experimental investigation. Qing Xu was responsible for the experimental design. Yixuan Zhang managed the dataset. Maggie M. He contributed to manuscript reviewing. Jian Zhang and Xiangjian He secured funding, administered and supervised the project, and were responsible for manuscript reviewing and editing.

\section*{Competing interests}\label{sec10}

The authors declare no competing interests.

\end{document}